\documentclass[runningheads]{llncs}
\usepackage[T1]{fontenc}
\usepackage{amsmath}
\usepackage{amssymb}
\usepackage{ bbold }
\usepackage{enumitem}
\usepackage{bbm}
\usepackage[font=small,skip=0pt]{caption}

\usepackage{graphicx}
\usepackage{color}

\newcommand{\norm}[1]{||#1||}

\begin{document}
\title{Weakly-Supervised Surgical Phase Recognition}
%
%

\author{Roy Hirsch\inst{1} \and Regev Cohen\inst{1} \and Mathilde Caron\inst{2}
\and Tomer Golany\inst{1} \and Daniel Freedman\inst{1} \and Ehud Rivlin\inst{1}}

\institute{Verily AI \and Google Research}

%
%
\maketitle              
\begin{abstract}
A key element of computer-assisted surgery systems is phase recognition of surgical videos. Existing phase recognition algorithms require frame-wise annotation of a large number of videos, which is time and money consuming. In this work we join concepts of graph segmentation with self-supervised learning to derive a random-walk solution for per-frame phase prediction. Furthermore, we utilize within our method two forms of weak supervision: sparse timestamps or few-shot learning. The proposed algorithm enjoys low complexity and can operate in low-data regimes. We validate our method by running experiments with the public Cholec80 dataset of laparoscopic cholecystectomy videos, demonstrating promising performance in multiple setups.

\end{abstract}
\section{Introduction}
\vspace{-5pt}
Automated phase recognition of laparoscopic videos plays an important role in computer-assisted surgery systems~\cite{ref_article_cleary}. Specifically, it allows for scheduling and monitoring surgical procedures, providing automatic offline analysis, and it can be used as a performance metric for novice surgeons. Recent deep learning solutions have shown unprecedented results, yet, such approaches typically require large numbers of labeled videos~\cite{ref_article_Cholec80,ref_article_tecno,ref_article_opera}, which are expensive and time-consuming to gather. In this preliminary work we introduce a simple yet effective solution for surgical phase recognition. We derive a method based on graph segmentation with weakly-supervised regularization priors. Our initial results show that it can handle data with limited annotation while providing promising performance.    

\vspace{-5pt}
\section{Problem Definition}
\vspace{-5pt}
Consider a surgical video $V=\{I_t\}_{t=0}^{T-1}$ of $T$ frames where $I_t\in\mathbb{R}^{A\times B\times C}$ denotes the image frame at time $t$. The video is comprised of $S\ll T$ consecutive temporal segments (phases), namely, there exists a partition $\Pi(V)=\{(I_t, y_t)\}_{t=0}^{T-1}$ that assigns each frame $I_t$ with a single phase label $y_t\in\{0,1,...,S-1\}$. Given the unlabeled test video $V$, the task of surgical phase recognition is concerned with devising an automatic system $\mathcal{M}(V)=\{(I_t, \hat{y}_t)\}_{t=0}^{T-1}$ to predict the frame-wise phase labels.
To that end, we assume a limited dataset $\mathcal{D}=\{V_n\}_{n=0}^{N-1}$ of $N$ surgical videos is at hand for training. In addition, we consider an offline setting where the complete test video is available at inference time.


\section{Proposed Method}
\vspace{-5pt}
Here we introduce our phase recognition solution by first translating it into a graph segmentation problem. We then describe our solution as a minimization problem that incorporates a weakly-supervised prior of one of two forms: (1)\;\textit{Sparse timestamp supervision} where only a small number of random frames within each phase are labeled at inference. (2)\;\textit{Few-shot supervision} where we use a small labeled dataset $\mathcal{D}$ to derive a spatial-temporal prior for the problem.

\vspace{-8pt}
\subsubsection{Graph Formulation and Random Walk Segmentation}
We start by reformulating the phase recognition task as 
a graph segmentation problem. Namely, for any given test video $V=\{I_t\}$ we define a sparse undirected weighted graph $G=(V,E, W)$ whose vertices are the video frames and edges are $E\triangleq \{e_{t,t+1}\;|\; t\in[0,1,...,T-2]\}$. Each edge $e_{i,j}\in E\subseteq V\times V$ spans two frames $I_i$ and $I_j$ and is assigned with a positive weight $w_{ij}\in W$, modeling the similarity between the two frames. We define the \textit{degree} of a vertex $I_i$ as $d_i\triangleq\sum_j w_{ij}$ where the sum is over all edges $e_{i,j}$ incident on $I_i$. 
The edge weights are given by $w_{ij}\triangleq\exp\Big(-\beta\frac{f_i^T f_j}{\|f_i\| \|f_j\|}\Big)$
where $f_i\triangleq\mathcal{F}(I_i)\in\mathbb{R}^M$ are the features of $I_i$ extracted by a model $\mathcal{F}$ pre-trained using self-supervised learning. Given the weighted graph, our goal is to label each vertex $I_i$ with a single phase $\hat{y}_i\in\{0,1,...S-1\}$. 

Let the vector $x^s\in\mathbb{R}^T$ hold the probabilities $x_t^s$ that the frame $I_t$ belongs to phase $s$, implying that $\sum_s x_t^s=1$ for any time-point $t\in\{0,1,...,T-1\}.$  Motivated by the random walk approach for image segmentation~\cite{ref_article_grady}, we formulate the following per-phase minimization using the graph Laplacian matrix $L$~\cite{ref_article_laplacian}:
\begin{equation}
\small
    \underset{x^s}{\min}\quad x^{sT}Lx^s+\gamma\norm{x^s-z^s}^2,\quad
    L_{ij} = 
        \begin{cases}
            d_i & i = j, \\
            -w_{ij} & e_{i,j}\in E, \\ 
            0 & \text{otherwise}.
        \end{cases}
    \label{eq:minimization}
\end{equation}
The first term encourages smoothness in the probability vector $x^s$; in the sparse timestamp setting, it leads to a diffusion of these timestamps.  The second term encourages the probability vector to be close to the prior $z^s$.  When taken together, we are guaranteed stable and unique solutions.  In particular, denoting
the identity matrix by $\mathcal{I}$, the solutions of (\ref{eq:minimization}) are readily given\footnote{\small In practice the solutions may not be probability vectors. Yet, this can be resolved by adding to all the solutions the constant $\mu = \frac{1}{S}\left(\mathbf{1} - \sum_{s=0}^{S-1} x^s\right)$.} by solving a sparse positive-definite system of linear equations $ \Big(L+\gamma \mathcal{I}\Big)x^s=\gamma z^s$ per phase.
Finally, our predictions are $\hat{y}_t =\underset{s}{\arg\max}\;x_t^s.$

Below, we complete our method and describe two distinct weakly-supervised approaches for modeling the prior vectors $\{z^s\}_{s=0}^{S-1}$. 

\begin{figure}[htb]
\centering
    \includegraphics[height=2cm, width=0.9\textwidth]{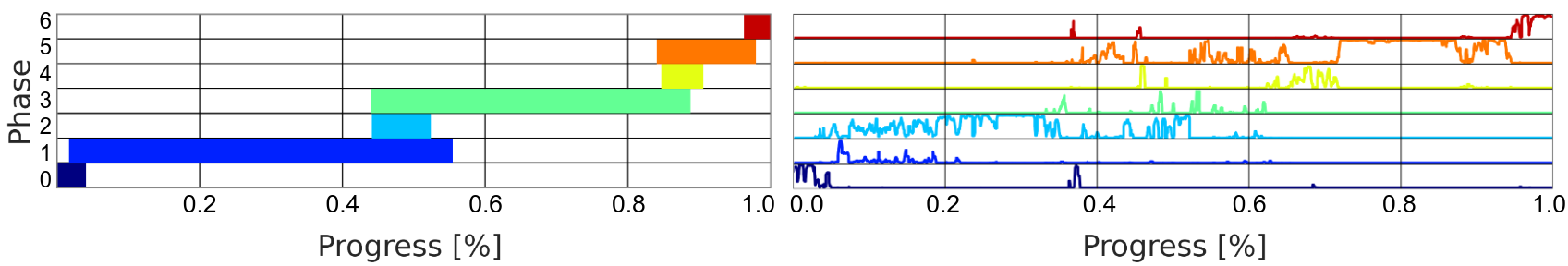}
    \caption{\textbf{Prior Visualization}: (left) binary temporal priors, and (right) spatial priors. The colors corresponds to the seven surgical phases of Cholec80 dataset.}
    \label{fig_priors}
\end{figure}
\subsubsection{Sparse Timestamp Supervision}
In this scenario we assume that for each phase of the test video a small, random and unique set $L^s=\{I_t\;|\:y_t\in s\}$ of $K\ll T$ labeled frames (timestamps) is given. Equipped with those labels, we define a sparse prior vector $z^s \in \mathbb{R}^T$  for each phase, whose entries are $z_t^s\triangleq\mathbb{1}(I_t\in L^s)$ where $\mathbb{1}(\cdot)$ is the indicator function. We then plug the above vectors into (\ref{eq:minimization}), incorporating timestamp supervision into our solution.  This approach enjoys the benefit of requiring only a handful of labels, thus saving annotation time and cost. Furthermore, assuming the feature-extractor $\mathcal{F}$ is given, no additional videos nor training are required.

\vspace{-3mm}

\subsubsection{Few-Shot Supervision}
In contrast to the above, here we assume no labels of the test video are provided. Instead, we consider a small dataset $\mathcal{D}$ of fully-labeled videos to perform few-shot learning. One direct approach is to train a deep learning model with the annotated dataset; however, we adopt a simpler approach which requires much quicker, more basic training. We define a spatial-temporal vector $z_t^s=u_t^s \cdot v_t^s$ as the product of the following elements (see Fig.~\ref{fig_priors}):
\begin{enumerate}
 \item \textbf{Spatial prior} We use the model $\mathcal{F}$ to extract frame-wise features for all videos in $\mathcal{D}$. Then, we group the feature vectors into $S$ clusters by their labels. We model the vectors within each cluster $C^s$ as samples of a multivariate Gaussian distribution and produce its density function $p_s(\cdot\,;\mu^s,\Sigma^s)$ where $\mu^s$ and $\Sigma^s$ are unbiased estimators of the distribution mean and covariance respectively. At inference, we define our spatial prior as the likelihood $u_t^s\triangleq p_s(f_t;\mu_s,\Sigma_s)$ where $f_t$ are the features of $I_t\in V$.

\item \textbf{Temporal prior} For any $V_n\in \mathcal{D}$ we construct a 2D histogram $H_n$ according to the frame time-points and labels where the number $N_x$ of time-bins equals the length of the shortest video and the number $N_y$ of phase-bins is $S$. Next, we average the $N$ histograms and then normalize each time-bin separately to yield $H$ such that $H_i^s$ holds the probability of frames at the time-bin $b_i$ to belong to phase $s$ and $\sum_s H_i^s=1$. At inference, we quantize the video time-points into $N_y$ equal time-bins as in $H$. Then, we build a binary temporal prior $v^s\in\mathbb{R}^T$ whose entries are given by $v_t^s=\mathbb{1}\big(H_i^s\geq\alpha^s\big)\cdot\mathbb{1}\big(t\in b_i\big)$. Namely, for any phase $s$ and time-point $t$ we obtain a time-bin $t\in b_i$ and the corresponding value $H_i^s$ and binarize it 
according to a threshold $\alpha^s>0$.
\end{enumerate}


\vspace{-5mm}

\section{Experiments} 
\vspace{-8pt}
\subsubsection{Setup}
We validate our approach using the Cholec80 dataset~\cite{ref_article_Cholec80} which contains 80 videos of laparoscopic cholecystectomy surgeries. Each video includes seven phases and is annotated by expert surgeons. We adopt the original data splits of 40/8/32 videos for training, validation and test respectively. As our feature extractor we use ViT-S, pre-trained by DINO~\cite{ref_article_dino} over ImageNet~\cite{ref_article_imagenet}, and we fine-tune it on the Cholec80 train fold using DINO which requires no labels. We train the model for 5000 steps using a batch size of 1024, the AdamW optimizer~\cite{ref_article_adamw} with a learning rate of $0.002$ and a linear scaling rule~\cite{ref_article_linear_lr}. Given the feature-extractor, we use the held-out Cholec80 validation set to tune our method hyperparameters by performing a grid search within the following ranges: $\beta \in [1, 10]$, $\gamma \in [1e^{-2}, 1e^{-4}]$ and $\alpha^s=\alpha\cdot\underset{i}{\max}\,H_i^s$ where $\alpha\in [0.4, 0.6]$. 

Finally, we evaluate our performance using the test set and we follow previous works to report the \textit{mean frame-wise accuracy}~\cite{ref_article_Cholec80,ref_article_tecno} and \textit{segmental F1 score}~\cite{ref_article_temp_conv} with overlapping thresholds of $10\%/25\%/50\%$.
While the former is a widely-used metric, we consider the latter as our key metric since it is treats all phases equally, independent of their duration, and it does not penalize for minor temporal shifts between the predictions and ground truth. 

\vspace{-12pt}
\subsubsection{Results and Discussion}
\begin{figure}[htb]
\centering
    \includegraphics[height = 4.5cm, width=0.8\textwidth]{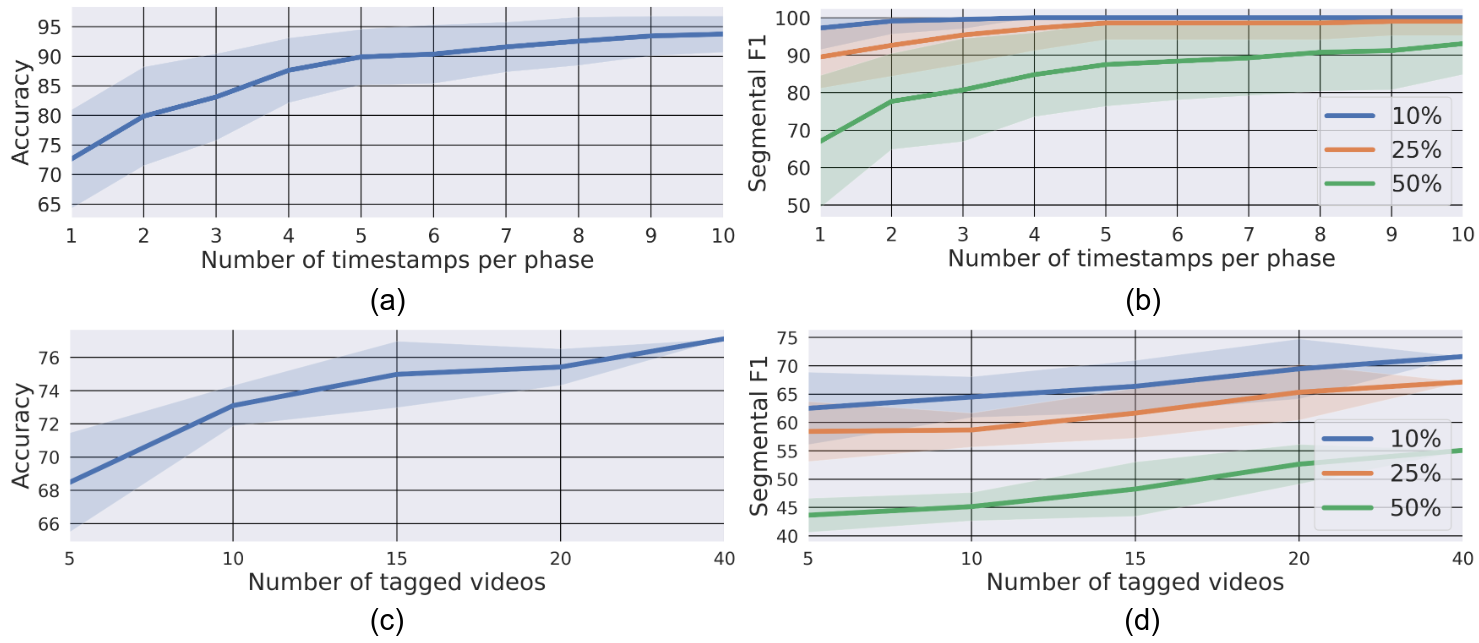}
    \caption{\textbf{Quantitative Result}: (top) sparse timestamp supervision, and (bottom) few-shot supervision. Results are averaged over 40 videos of Cholec80 test set.}
    \label{fig_mets}
\end{figure}

First we consider the \textit{sparse timestamp supervision} scenario where $K\in[1,2,...10]$. We run each experiment five times with random seeds and report the mean and standard deviation of each metric. The results in Fig.~\ref{fig_mets} show that even for single timestamp ($K=1$) the model achieves adequate performance of $72.6\%$ accuracy and $97.24$ F1@10. As expected, the performance improves as the number of timestamps increases, $90\%$ accuracy and $100$ F1@10 for the small number of 5 timestamps. 
Next, we consider \textit{few-shot supervision} where we study  different sizes of $\mathcal{D}$ ($N\in[5, 10, 15, 20, 40]$). Fig.~\ref{fig_mets} shows that our method displays promising results for the relatively small number of 15 annotated videos, reaching $75\%$ accuracy and $65$ F1@10.

To summarize, we have presented a simple yet effective algorithm for surgical phase recognition based on graph random walks guided by deep features and weak supervision.
Our experiments with laparoscopic cholecystectomy videos show that our method can cope with limited data 
-- either few timestamps or few annotated videos -- 
while attaining good performance on surgical phase detection. This initial work lays the groundwork for self-supervised (or unsupervised) phase detection methods which require little or no annotated data. 

\vspace{-15pt}
\begin{figure}[htb]
\centering
    \includegraphics[height = 1cm, width=0.8\textwidth]{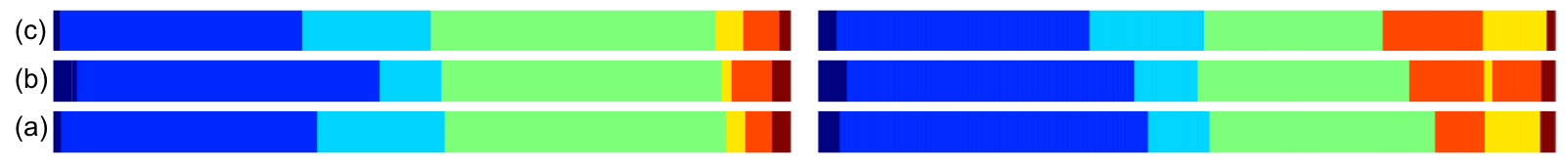}
    \caption{\textbf{Qualitative Results:} (a) ground true, (b) few-shot supervision and (c) sparse timestamp supervision. Results obtained using 2 test videos of Cholec80 (left and right).}
    \label{fig_samples}
\end{figure}

\vspace{-20pt}

\clearpage

\begin{thebibliography}{8}
\bibitem{ref_article_cleary}
Cleary, K., Chung, H.Y. and Mun, S.K.\textit{ OR2020 workshop overview: operating room of the future.} International Congress Series, Vol. 1268, pp. 847-852, Elsevier, June 2004.

\bibitem{ref_article_Cholec80}
Twinanda, A.P., Shehata, S., Mutter, D., Marescaux, J., De Mathelin, M. and Padoy, N. \textit{EndoNet: a deep architecture for recognition tasks on laparoscopic videos}. IEEE Transactions on Medical Imaging, 36(1), pp.86-97, 2016.

\bibitem{ref_article_tecno}
Czempiel, T., Paschali, M., Keicher, M., Simson, W., Feussner, H., Kim, S.T. and Navab, N. \textit{Tecno: Surgical phase recognition with multi-stage temporal convolutional networks.} International Conference on Medical Image Computing and Computer-Assisted Intervention (pp. 343-352) Springer, Cham.  October 2020.

\bibitem{ref_article_opera}
Czempiel, T., Paschali, M., Ostler, D., Kim, S.T., Busam, B. and Navab, N., 2021, September. \textit{Opera: Attention-regularized transformers for surgical phase recognition.} International Conference on Medical Image Computing and Computer-Assisted Intervention (pp. 604-614). Springer, Cham. September 2021.

\bibitem{ref_article_laplacian}
Merris, R., 1994. \textit{Laplacian matrices of graphs: a survey.} Linear Algebra and Its Applications, 197 (1994), pp.143-176.

\bibitem{ref_article_temp_conv}
Lea, C., Flynn, M.D., Vidal, R., Reiter, A. and Hager, G.D. \textit{Temporal convolutional networks for action segmentation and detection.} Proceedings of the IEEE Conference on Computer Vision and Pattern Recognition (pp. 156-165), 2017.

\bibitem{ref_article_temp_cvpr21_timestamp}
Li, Z., Abu Farha, Y. and Gall, J. \textit{Temporal action segmentation from timestamp supervision.} Proceedings of the IEEE/CVF Conference on Computer Vision and Pattern Recognition (pp. 8365-8374), 2021.

\bibitem{ref_article_dino}
Caron, M., Touvron, H., Misra, I., Jégou, H., Mairal, J., Bojanowski, P. and Joulin, A. \textit{Emerging properties in self-supervised vision transformers.} Proceedings of the IEEE/CVF International Conference on Computer Vision (pp. 9650-9660), 2021.

\bibitem{ref_article_imagenet}
Krizhevsky, A., Sutskever, I. and Hinton, G.E. \textit{Imagenet classification with deep convolutional neural networks.} Advances in Neural Information Processing Systems, 25, 2012.

\bibitem{ref_article_adamw}
Loshchilov, I. and Hutter, F., 2018. \textit{Fixing weight decay regularization in ADAM}. URL: https://openreview.net/pdf?id=rk6qdGgCZ.


\bibitem{ref_article_grady}
Grady, L. \textit{Multilabel random walker image segmentation using prior models.} IEEE Computer Society Conference on Computer vision and Pattern Recognition. Vol. 1, pp. 763-770, 2005.

\bibitem{ref_article_linear_lr}
Goyal, P., Dollár, P., Girshick, R., Noordhuis, P., Wesolowski, L., Kyrola, A., Tulloch, A., Jia, Y. and He, K. Accurate, large minibatch SGD: Training Imagenet in 1 hour. arXiv preprint arXiv:1706.02677.

\end{thebibliography}
%

\end{document}